\documentclass[letterpaper]{article} 
\usepackage[utf8]{inputenc}
\usepackage{aaai2026}
\usepackage{times}  
\usepackage{helvet}  
\usepackage{courier}  
\usepackage[hyphens]{url}  
\usepackage{graphicx} 
\usepackage{booktabs}
\usepackage{multirow}
\usepackage{algorithm}
\usepackage{algpseudocode}
\usepackage{amsmath}
\usepackage{amssymb}
\usepackage{mathtools}
\usepackage{tikz}
\usetikzlibrary{shapes.geometric, patterns, arrows.meta, positioning}
\usepackage{subcaption} 
\usepackage{natbib}  
\usepackage{caption} 
\usepackage{color}

\usetikzlibrary{arrows.meta}

\usepackage{newfloat}
\usepackage{listings}
\usepackage{comment}
\DeclareCaptionStyle{ruled}{labelfont=normalfont,labelsep=colon,strut=off} 
\floatstyle{ruled}
\newfloat{listing}{tb}{lst}{}
\floatname{listing}{Listing}
\usepackage{tikz}
\usepackage{caption}
\usetikzlibrary{shapes.geometric, patterns, arrows.meta, positioning}
\usepackage{subcaption} 
\title{Cooperative-ORCA$^*$: Real-Time Proactive Deadlock Avoidance for Continuous-Space Multi-Agent Navigation}
\author {
    Junfeng Wu\textsuperscript{\rm 1},
    Jiaqi Chen\equalcontrib\textsuperscript{\rm 1},
    Hongkun Lyu\equalcontrib\textsuperscript{\rm 1},
    Kevin Zheng\textsuperscript{\rm 1},
    Andy Li\textsuperscript{\rm 1}
}
\affiliations {
    \textsuperscript{\rm 1}Monash University\\
    \{jwuu0222, jche0604, hlyu0021\}@student.monash.edu\\
    \{Kevin.Zheng, andy.li\}@monash.edu
}

\usepackage{bibentry}
\usepackage{comment}
\begin{document}

\maketitle

\begin{abstract}
Multi-Agent Path Finding (MAPF) is a problem that requires computing collision-free paths for a set of agents from their start locations to designated goal locations.
The problem has broad applications in domains where teams of robots must operate in a coordinated manner.
ORCA* is a real time MAPF solver that assigns for each timestep a velocity for each agent.
Due to its real time nature, it is myopic to future deadlocks that result from current decisions.
ORCA*-MAPF attempts to remedy this limitation by introducing fallback mechanisms when deadlocks are detected. However, post hoc interventions often introduce significant flowtime overhead. In this paper, we introduce C-ORCA* and C-ORCA*-MAPF, continuous space MAPF algorithms that incorporate agents' entire spatial trajectory and their spatial dependencies to proactively prevent deadlocks from occurring, thus avoiding the high flowtime overhead associated with post hoc corrections in ORCA*-MAPF. The C-ORCA* family of algorithms significantly outperform previous state-of-the-art in terms of solve rate, runtime, and flowtime.   
\end{abstract}


\section{Introduction}
Multi-Agent Path Finding (MAPF) is the problem of planning collision-free paths for multiple agents from their respective start locations to their goal locations within a shared environment~\cite{stern2019}. This problem is ubiquitous in industrial and commercial settings, including warehouse robotics, delivery robots in restaurants, and assembly lines. With the continued push to bring AI, automation, and robotics into practice, the need to solve MAPF effectively has never been greater.

MAPF is classically modelled as a problem on a discrete graph or grid environment with discretised time. Agents 'jump' to direct neighbours from their current position for each time step. Researchers have produced numerous algorithms that solve this relaxed problem, typically involving some form of search; state-of-the-art solvers can find paths for about a million agents~\cite{advancing} in some cases. Needless to say, this relaxed model fails to capture many nuances of operating real-world robotic fleets. For one, physical robots traverse a continuous environment. Moreover, the motions of the robots are non-discrete, subject to hardware design and kino-dynamics. Classical MAPF solutions do not trivially transfer to the real world.   

Configuration Space MAPF (CS-MAPF)~\cite{choset2005configurationspace} extends the classical MAPF problem to continuous domains. In this paradigm, a \textit{configuration} typically denotes a set of valid positions occupied by the fleet of robots at a given time. CS-MAPF problems are commonly solved in an \textit{action space}, a reduced set of motion primitives that transition the system from one configuration to another~\cite{db-cbs,db-LaCAM, space-action1,spaceaction2}. Because robot kinematics and dynamics are implicitly captured in this space, the model more closely reflects real-world robotic behaviour. However, despite this increased fidelity, CS-MAPF solutions remain largely theoretical. Even when they are feasible, the post-processing or control strategies required for execution often lead to realised costs that significantly exceed the expected solution cost~\cite{yan2026advancing}.

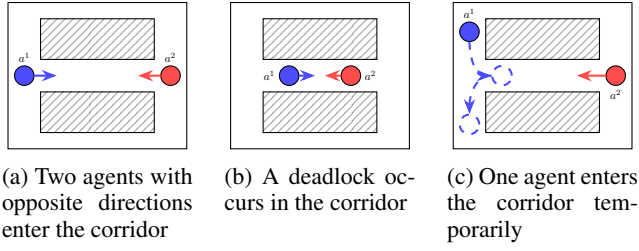
\begin{figure}[t]
    \centering
    \begin{subfigure}[t]{0.3\columnwidth}
        \centering
        \begin{tikzpicture}[scale=0.43, every node/.style={scale=0.43}]
      
            \draw[thin, black] (0,0) rectangle (5.5, 4.5);
         
            \fill[pattern=north east lines, pattern color=gray!70, draw=black] (1, 2.75) rectangle (4.5, 4.0);
            \fill[pattern=north east lines, pattern color=gray!70, draw=black] (1, 0.5)  rectangle (4.5, 1.75);

            \node[circle, fill=blue!70, draw=black, minimum size=0.6cm, inner sep=0pt, label=above:$a^1$] (a1) at (0.5, 2.25) {};
            \draw[-{Stealth[length=2mm]}, thick, blue!70] (a1) -- (1.5, 2.25);

            \node[circle, fill=red!70, draw=black, minimum size=0.6cm, inner sep=0pt, label=above:$a^2$] (a2) at (5.0, 2.25) {};
            \draw[-{Stealth[length=2mm]}, thick, red!70] (a2) -- (4.0, 2.25);
            
        \end{tikzpicture}
        \caption{Two agents with opposite directions enter the corridor}
        \label{fig: senario:enter}
    \end{subfigure}
    \hfill
    \begin{subfigure}[t]{0.3\columnwidth}
        \centering
        \begin{tikzpicture}[scale=0.43, every node/.style={scale=0.43}]
          
            \draw[thin, black] (0,0) rectangle (5.5, 4.5);

            \fill[pattern=north east lines, pattern color=gray!70, draw=black] (1, 2.75) rectangle (4.5, 4.0);
            \fill[pattern=north east lines, pattern color=gray!70, draw=black] (1, 0.5)  rectangle (4.5, 1.75);

            \node[circle, fill=blue!70, draw=black, minimum size=0.6cm, inner sep=0pt, label=left:$a^1$] (a1) at (1.8, 2.25) {};
            \node[circle, fill=red!70, draw=black, minimum size=0.6cm, inner sep=0pt, label=right:$a^2$] (a2) at (3.7, 2.25) {};

            \draw[-{Stealth[length=2mm]}, thick, blue!70] (a1) -- (2.6, 2.25);
            \draw[-{Stealth[length=2mm]}, thick, red!70] (a2) -- (2.9, 2.25);
            
        \end{tikzpicture}
        \caption{A deadlock occurs in the corridor}
        \label{fig: senario:ddl}
    \end{subfigure}
    \hfill
    \begin{subfigure}[t]{0.3\columnwidth}
        \centering
        \begin{tikzpicture}[scale=0.43, every node/.style={scale=0.43}]
            
            \draw[thin, black] (0,0) rectangle (5.5, 4.5);

            \fill[pattern=north east lines, pattern color=gray!70, draw=black] (1, 2.75) rectangle (4.5, 4.0);
            \fill[pattern=north east lines, pattern color=gray!70, draw=black] (1, 0.5)  rectangle (4.5, 1.75);
         
            \node[circle, fill=blue!70, draw=black, minimum size=0.6cm, inner sep=0pt, label=above:$a^1$] (a1top) at (0.5, 3.6) {};

            \node[circle, draw=blue!70, dashed, fill=white, thick, minimum size=0.6cm, inner sep=0pt] (a1mid) at (1.5, 2.25) {};

            \node[circle, draw=blue!70, dashed, fill=white, thick, minimum size=0.6cm, inner sep=0pt] (a1bot) at (0.5, 0.75) {};

            \node[circle, fill=red!70, draw=black, minimum size=0.6cm, inner sep=0pt, label=below:$a^2$] (a2) at (5.0, 2.25) {};

            \draw[-{Stealth[length=2mm]}, thick, blue!70, dashed] (a1top) to[out=270, in=180] (a1mid);
            
            \draw[-{Stealth[length=2mm]}, thick, blue!70,dashed] (a1mid) to[out=180, in=90] (a1bot);

            \draw[-{Stealth[length=2mm]}, thick, red!70] (a2) -- (3.8, 2.25);
            
        \end{tikzpicture}
        \caption{One agent enters the corridor temporarily}
        \label{fig: senario:temp enter}
    \end{subfigure}
    \caption{Different corridor situations}
    \label{fig: senario}
\end{figure}

Optimal Reciprocal Collision Avoidance (ORCA$^*$) \cite{orca-originial,orca*} is a decentralised strategy for guiding robotic fleets in real time. It operates on the velocities of the agents directly, which is continuous and closer to the execution of the robots.
Given an ideal velocity for each of the agents at the current timestep, ORCA$^*$ assigns a new velocity for each agent for the next time step that prevent agents from colliding with obstacles or each other.
ORCA$^*$ is an effective but purely reactive guidance strategy. As such, velocities computed at the current timestep may lead to future deadlocks. This is most evident in corridor-like situations. 
Fig \ref{fig: senario} shows an example of when two agents enter a long corridor from each side (Fig \ref{fig: senario:enter}). They deadlock when they meet in the center (Fig \ref{fig: senario:ddl}).
At the time of entrance, the agents were far apart, and their eventual encounter was not clear, as one agent could just be moving into the corridor temporarily to avoid other agents (Fig \ref{fig: senario:temp enter}).
ORCA$^*$-MAPF \cite{mapf-orca} addresses deadlock situations by introducing a fallback mechanism.
When an agent’s velocity remains near zero for a specified duration, i.e. deadlock, a centralised grid-based MAPF algorithm is invoked to replan paths.
However, this approach has two notable limitations.
First, the mechanism is only triggered after a deadlock has already occurred, causing agents to spend unnecessary time in stalled states before detection, which is inherently inefficient.
Second, the solution generated by the MAPF algorithm is discretised and subsequently post-processed into motions with fixed speeds (1 for movement and 0 for waiting) and restricted to the four cardinal directions.
This discretisation undermines the advantages of ORCA$^*$operating in a continuous space.

We introduce a new continuous space solver, Cooperative ORCA$^*$ (C-ORCA$^*$), which is capable of avoiding deadlocks before they occur.
As a pre-processing step, C-ORCA$^*$ obtains a guidance MAPF solution from a classical MAPF planner, such as MAPF-LNS \cite{mapf-lns2,jcai2021p568}, as a global and long-term heuristic for the agents' trajectory.
These paths are converted into way-points, i.e. intermediate goal locations, via a string-pulling algorithm.
We also identify dependencies between agents during pre-processing.
During execution, agents are instructed to move from one way-point to the next until the goal location. A modified version of ORCA$^*$ is run for each agent per iteration, taking into account their preferred velocity as well as spatial dependency information to compute the actual velocity.
In C-ORCA$^*$, agents may able to wait proactively to prevent future potential deadlock and introduce a strong cooperative behaviour. We also propose C-ORCA$^*$-MAPF, a variant of C-ORCA$^*$ that also inherits the fallback behaviour of ORCA$^*$-MAPF. The C-ORCA$^*$ family of algorithms significantly outperform previous state-of-the-art in terms of solve rate, runtime, and flowtime.  

\section{Problem Statement}
The Configuration Space MAPF~\cite{choset2005configurationspace} problem takes a tuple of an environment and a set of agents, $\langle \chi,A\rangle$ as an input.
The environment is partitioned into two regions, $\chi = \chi_{free} \cup \chi_{obs}$, where $\chi_{free}$ denotes the traversable free space and $\chi_{obs}$ denotes static obstacles with which agents must not collide with.
Agents $A=\{a^1,...,a^n\}$ is a set of $n$ agents, where each agent $a^i$ has an initial configuration and goal configuration. 
CS-MAPF is required to compute a set of collision-free trajectories $\Pi=\{\pi^1,...,\pi^n\}$, one for each agent, from their initial configuration to the goal configuration.

The configuration space considered in this paper is a two-dimensional rectangular metric space $\chi \in \mathbb{R}^2$. Thus we will use a grid world $G$ containing a set of points $V$ to express the traversability of the environment. Each point $v_i \in V$ in $G$ is a tuple of $\langle x_i,y_i,b_i \rangle$, where $b_i$ is a boolean value $\{0,1\}$ to represent whether each point is traversable or not.
Therefore, $\chi_{obs}$ is the non-traversable area in $G$.
Each agent is modelled as a disc of radius $r$, capable of moving in any direction with a maximum speed of $1$.
The position of agent $a^i$ at time $t$ is represented in the Cartesian coordinate system as $\boldsymbol{p}^i_t = (x^i_t, y^i_t)$.
We consider two types of collisions under this model:
\begin{itemize}
    \item \textbf{Agent–obstacle collision:} occurs when the minimum Euclidean distance between the agent’s centre and the boundary of any obstacle is less than $r$.
    \item \textbf{Inter-agent collision:} occurs when the Euclidean distance between two agents' center is less than $2r$.
\end{itemize}
A problem instance is considered \emph{solved} if all agents can safely park at their goal configuration.
The cost of each agent is the earliest time at which it safely reaches its goal configuration.
The objective is to minimise the flowtime, i.e., the sum of all agents’ costs.

\section{Related Work}
In this section, we review existing strategies that `directly' plan for the continuous-space (CS-MAPF): state lattice planning and the ORCA$^*$ family of algorithms.
We acknowledge a broader body of related work \cite{MAPF-POST,switch-ADG,PIE} that seeks to bridge the gap between discrete MAPF solutions and execution in physical environments via grid plan adaptation. However, these methods are primarily concerned with ensuring the faithful execution of a precomputed MAPF plan, typically preserving the structure and guarantees of the underlying grid-based solution.
In contrast, this line of work prioritises decentralised decision-making and responsiveness in continuous space, tolerating deviations from the trajectories prescribed by a discrete MAPF solution.
The line of study on Any-Angle planning methods such as AA-SIPP and AA-CBS appear related, but they operate on fully connected graphs and do not explicitly model motion constraints, such as curvature or kinematic feasibility.
Moreover, these approaches typically assume uniform, constant velocities across agents, limiting their expressiveness in continuous domains. Consequently, they are not considered in our discussion.


\subsection{State Lattice Planning}
State Lattice Planning is a motion planning technique that discretises a robot’s continuous state space into a structured lattice of feasible states.
Each agent is associated with a set of motion primitives, which define how the agent can transition from one state to another.
Unlike classical grid-based planning, which only considers positional discretisation, motion primitives incorporate the system’s kinematic and dynamic constraints, ensuring that generated paths are feasible with respect to the agent’s motion capabilities.
Recent research in MAPF has explored incorporating the state lattice model into MAPF, as demonstrated by db-CBS~\cite{db-cbs} and db-LaCAM~\cite{db-LaCAM}.
Both algorithms extend classical MAPF solvers by replacing the single-agent planning component with a state-lattice–based solver.
db-CBS builds on Conflict-Based Search~\cite{2015-cbs}, an optimal MAPF algorithm, and provides probabilistic completeness as well as asymptotic optimality guarantees.
However, its scalability is severely limited: it can solve only up to eight agents in relatively simple environments, compared to classical CBS in standard MAPF, which can handle hundreds of agents.
Similarly, db-LaCAM integrates the LaCAM solver~\cite{Okumura_2023}, a highly scalable but sub-optimal MAPF algorithm.
While LaCAM can solve tens of thousands of agents in classical MAPF, db-LaCAM using motion primitives is restricted to around 50 agents.
This contrast demonstrates that incorporating motion primitives introduces significant computational challenges.
Moreover, state-lattice–based approaches are inherently less responsive to execution uncertainty, such as deviations from prescribed motion primitives, limiting their adaptability in practical settings or requiring post-processing during execution.
\subsection{Collision Avoidance in Continuous Space}                                                                                                           
In contrast to cooperative search methods for solving CS-MAPF, Optimal Reciprocal Collision Avoidance (ORCA$^*$) \cite{orca-originial,orca-a*} is a decentralised collision avoidance algorithm that operates directly in the agent’s velocity space.
ORCA$^*$ builds on the concept of Velocity Obstacles (VO) \cite{rvo}, where each agent identifies a set of forbidden velocities that would result in collisions with neighbouring agents within a short time horizon.
By selecting velocities outside this set, agents can maintain collision-free motion under the assumption that all other agents adhere to the same reciprocal rules. Algorithm \ref{algo: c-orca}'s non-highlighted lines exemplify ORCA$^*$'s control procedure, each agent continuously surveys its neighbours' trajectories (lines 4, 5) and applies an optimal and collision-free velocity upon itself (lines 9, 10) until termination.

ORCA$^*$ is distributed and computationally efficient, making it suitable for real-time applications with large numbers of agents, such as crowd simulation and autonomous robotics. In benchmarks, ORCA$^*$ performs well when there is a large open area.
However, as a purely reactive method, it does not perform any sort of global planning or coordination and thus can become easily trapped in local deadlocks given constrained environments. The most devastating way this surfaces is in cases involving narrow corridors where traffic must flow bidirectionally \cite{limit2}. \textit{gap1-64-64} is a map with two large open areas and a wall with a unit sized gap. Observe in Figure~\ref{fig:experiment} that the success rate drops sharply for ORCA$^*$, falling below 50\% at roughly 40 agents; they pile up and block the gap in the wall.

As an attempt to imbue some sense of coordination, \citet{mapf-orca} proposed ORCA$^*$-MAPF, which invokes a classical MAPF solver as a fallback: when a group of agents’ velocities remain low for a period of time, indicating a deadlock, the solver generates a discrete solution to `untie' the agents.
ORCA$^*$ resumes execution once the agents are freed.
ORCA$^*$-MAPF addresses the weaknesses of ORCA$^*$ in theory but only achieves limited improvements over ORCA$^*$ empirically. This is for a few reasons. First, the fallback mechanism activates only after a deadlock occurs, introducing unnecessary delays. Furthermore, the MAPF solution is discrete and sometimes infeasible for agents to execute, especially in post-deadlock configurations where agents may be densely packed. More generally, the discrete nature of the fallback mechanism undermines the continuous-space nature of ORCA$^*$.

\vspace{1\baselineskip} 
\section{C-ORCA$^*$}
\begin{algorithm}[t]
\caption{C-ORCA*}
\label{algo: c-orca}
\begin{algorithmic}[1]
    \State \textbf{Input:} $\langle G, A\rangle$
    
    \State \textcolor{red}{Pre-process$(\langle G, A\rangle)$}
    
    \While{$\neg Terminate$}
        \State $A \gets \text{TakeObservation}()$ \Comment{if in practical setting}
        \State $\mathcal{A} \gets \text{GetNeighbors}(A)$
        \State \textcolor{red}{$WP' \gets$ GetWaypoint($A,WP,\mathcal{A}$)}
        \State \textcolor{red}{CheckDependency($A,\mathcal{C},\mathcal{C}_{dep}$)}
        \State \textcolor{red}{$V^{pref} \gets$ PreferredVelocity($A,WP'$)}
        \State $V^{new} \gets \text{ComputeVelocity}(\mathcal{A},V^{pref})$ 
        \State ApplyControl($V^{new}$)
        

        
    \EndWhile
\end{algorithmic}
\end{algorithm}

\begin{algorithm}[t]
\caption{Pre-Process ($G, A$)}
\label{algo: pre-process}
\begin{algorithmic}[1]
    
    \State $\Pi' \gets \text{MAPFPlanner}(G)$
    \State $\mathcal{C} \gets \text{IdentifyCorridors}(G)$
    \State $\mathcal{C}_{dep} \gets \text{BuildDependency}(\Pi', \mathcal{C})$ 
    \State $WP \gets \text{StringPulling}(\Pi')$ 

\end{algorithmic}
\end{algorithm}
In this section, we present our approach to CS-MAPF, Cooperative ORCA$^*$ (C-ORCA$^*$), which leverages the strengths of both cooperative search methods and local collision avoidance to proactively avoid deadlocks. C-ORCA$^*$ is a real-time technique; thus, planning is done during execution and can react to the execution-time uncertainties, such as drifting or delay. Like ORCA$^*$, C-ORCA$^*$ scales well to larger fleets. Yet, it effectively handles challenging corridor scenarios --- situations in which the original ORCA$^*$ algorithm struggled greatly.

Algorithm \ref{algo: c-orca} gives an overview of the technique and show how each component is connected. The highlighted lines indicate modifications made to classic ORCA$^*$. C-ORCA$^*$'s pre-processing step (line 2 in Algorithm~\ref{algo: c-orca} and described in Algorithm~\ref{algo: pre-process}) computes the necessary information for future planning, namely a sequence of waypoints for each agent, corridors, and potential dependencies for agents during execution.
During planning and execution, C-ORCA$^*$ takes this information and combines it with real-time observations to continuously calculate a velocity for each agent. Algorithm~\ref{algo: c-orca}  lines 6--8 outline this process, including neighbour and way-point identification, dependency check, and calculation of a preferred and final velocity. C-ORCA$^*$ terminates once the goal configuration is achieved.

\subsection{Pre-Processing}
The pre-processing stage begins by converting the problem environment into a grid-based one, allowing us to leverage classical MAPF solvers to produce guidance paths for real-time planning.
All corridor-like structures within the map are identified. Computations in corridor detection are also performed within this grid-based setting. 
Then, for each corridor, we identify all the relative dependencies among agents based on their guidance path.
Finally, we convert the grid guidance path for each agent into a sequence of way-points that direct the agent to the goal location making use of a string-pulling post-process technique.

\paragraph{Guidance Path Generation:} The guidance path is a classical MAPF solution for a grid discretisation of the problem instance. Any classical MAPF solver can be invoked to generate the guidance path.

\paragraph{Corridor Detection:} Generally speaking, a corridor is a narrow region that only allows one agent to traverse through.
Formulating the problem instance as classical MAPF, where the environment is a 4-connected grid $G=\langle V, E \rangle$, and $V$ is the set of freely traversable cells in the grid, $E$ is a set of edges that connect each cell to its four adjacent neighbouring cells.
We define a cell $v\in V$ to be a corridor cell if the degree of the vertex is less than or equal to 2, i.e. $deg(v)\leq2$.
We consider $v_i,v_j\in V$ to belong to the same corridor if there exists an ordered list of vertices $V'=[v_i,v_1,...,v_k,v_j]$ which connects $v_i$ and $v_j$, and there exists an edge $e' \in E$ for every two consecutive vertices $v_i',v_j' \in V'$, and every vertex is a corridor cell, i.e., $deg(v')\leq 2~|~v' \in V'$.
For each corridor cell $v'$, the corridor $c$ that it belongs to is the largest ordered list $V'=[v_1,...,v',...,v_n]$ which contains $v'$. The first $v_1$ and last $v_n$ vertices in $V'$ are considered the two entrances of $V'$.
We use $\mathcal{C}$ to denote the set of all the corridors in $G$. $c^{e_1}_i$ and $c^{e_2}_i$ denote the two entrance vertices of corridor $c_i$.

Corridors $\mathcal{C}$ in $G$ are identified via a linear scan of the grid, commencing from the top-left to the bottom-right, evaluating the degree of each vertex. 
When a vertex $v$ is identified as a corridor cell, it is treated as one entrance of a corridor. If $\deg(v) = 2$, a depth-first search is initiated from $v$ toward its unscanned neighbour, proceeding until a vertex with degree 1 or greater than 2 is encountered.
All intermediate vertices are marked as scanned and classified as part of the corridor.
The final vertex along this path with degree $\leq 2$ is designated as the opposite entrance of the corridor.
If $\deg(v) = 1$, then it's a single-cell corridor.
The scan then resumes, skipping all previously scanned vertices.
This procedure identifies and groups all corridor cells in $O(N)$ where $N$ is the number of vertices $|V|$ in $G$.

\paragraph{Dependency Detection}

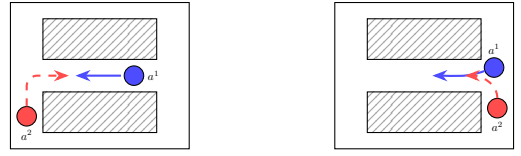
\begin{figure}[t]
    \centering
    \begin{subfigure}[t]{0.49\columnwidth}
        \centering
        \begin{tikzpicture}[scale=0.43, every node/.style={scale=0.43}]

    \draw[thin, black] (0,0) rectangle (5.5, 4.5);

    \fill[pattern=north east lines, pattern color=gray!70, draw=black] (1, 2.75) rectangle (4.5, 4.0);
    \fill[pattern=north east lines, pattern color=gray!70, draw=black] (1, 0.5)  rectangle (4.5, 1.75);

    \node[circle, fill=red!70, draw=black, minimum size=0.6cm, inner sep=0pt, label=below:$a^2$] (a2) at (0.5, 1.0) {};

    \node[circle, fill=blue!70, draw=black, minimum size=0.6cm, inner sep=0pt, label=right:$a^1$] (a1) at (3.8, 2.25) {};

    \draw[-{Stealth[length=2mm]}, thick, blue!70] (3.4, 2.25) -- (2.0, 2.25);

    \draw[-{Stealth[length=2mm]}, thick, red!70, dashed, rounded corners] (a2) |- (1.8, 2.25);
    
\end{tikzpicture}
        \caption{$a^1$ enters the corridor first, thus $a^2$ waits until $a^1$ leaves}
        \label{fig: cross-corridor}
    \end{subfigure}
    \hfill
    \begin{subfigure}[t]{0.49\columnwidth}
        \centering
        \begin{tikzpicture}[scale=0.43, every node/.style={scale=0.43}]

    \draw[thin, black] (0,0) rectangle (5.5, 4.5);
   
    \fill[pattern=north east lines, pattern color=gray!70, draw=black] (1, 2.75) rectangle (4.5, 4.0);
    \fill[pattern=north east lines, pattern color=gray!70, draw=black] (1, 0.5)  rectangle (4.5, 1.75);

    \node[circle, fill=red!70, draw=black, minimum size=0.6cm, inner sep=0pt, label=below:$a^2$] (a2) at (5.0, 1.25) {};
    \node[circle, fill=blue!70, draw=black, minimum size=0.6cm, inner sep=0pt, label=above:$a^1$] (a1) at (4.9, 2.5) {};

    \draw[-{Stealth[length=2mm]}, thick, blue!70] (a1) to[out=200, in=0] (3.0, 2.25);
    \draw[-{Stealth[length=2mm]},thick, red!70, , dashed]  (a2) to[out=90, in=0] (4.0, 2.25);
    
\end{tikzpicture}
        \caption{$a^1$ arrives the intersection to corridor first, thus $a^2$ wait until $a^1$ fully enters corridor}
        \label{fig: enter-corridor}
    \end{subfigure}
    \caption{Different dependency situations}
    \label{fig: dependency}
\end{figure}

The computed guidance path is a classical MAPF solution; each agent's plan $\pi^i$ is an ordered list of vertices $\pi^i=[v_1^i,...,v^i_k]$.
For each corridor $c_i \in \mathcal{C}$, we compute two sets of agents $A_{i_1}$ and $A_{i_2}$.
An agent $a^i \in A_{i_1}$ (resp. $a^j \in A_{i_2}$) must contain the entirety of the corridor $c_i$ in its plan $\pi^i$ (resp. $\pi^j$),
It enters from $c^{e_1}_i$ (resp. $c^{e_2}_i$) and leaves from $c^{e_2}_i$ (resp. $c^{e_1}_i$). 

We consider two types of dependency here:
\begin{itemize}
    \item \textbf{Cross-corridor dependency}: Any agent $a^i \in A_{i_1}$ (resp. $a^j \in A_{i_2}$) has a dependency relationship with all agent $a^j \in A_{i_2}$ (resp. $a^i \in A_{i_1}$).
    We use $\xLeftrightarrow{c_i}$ to denote the cross-corridor dependency between two agents or two sets of agents over corridor $c_i$, e.g., $a^i \xLeftrightarrow{c_i} a^j$ or $A_{i_1} \xLeftrightarrow{c_i} A_{i_2}$.
    The cross-corridor dependency will force one agent to wait if the another agent enters the corridor from the opposite direction earlier, as demonstrated in Fig. \ref{fig: cross-corridor}. 
    This proactively prevents the deadlock situation from happening.
    \item \textbf{Enter-corridor dependency}: An enter-corridor dependency exists for any two agents $a^i,a^j \in A_{i_1}$ (resp. $a^i,a^j \in A_{i_2}$).
    We use $a^i \overset{c_i}{\approx} a^j$ to denote the enter-corridor dependency between $a^i$ and $a^j$ for corridor $c^i$.
    The enter-corridor dependency will force one agent to wait until the other agent fully enters the corridor if the other agent arrives the corridor earlier, as demonstrated in Fig. \ref{fig: cross-corridor}. 
    This proactively prevents multiple agents from competing to enter the corridor at the same time.
\end{itemize}

Recall the guidance path is computed from a discrete relaxation of the problem. If $a^i$ arrives earlier than $a^j$ in the guidance path, it does not necessarily mean that it will be the case in execution. 
Thus, the dependency was built ignoring the time dimension, i.e., if $a^1$ leaves $c^{e_1}_i$ at $t_i$ in the guidance path, and $a^2$ enters $c^{e_1}_i$ at $t_2$, where $t_1 \ll t_2$.
$a^1 \xLeftrightarrow{c_i} a^2$ is still considered, similarly for enter-corridor dependency.

This means the dependency between $a^i$ and $a^j$ is an association rather than an ordering. That is to say, given $a^1 \xLeftrightarrow{c_i} a^2$, the traversal order through $c_i$ is determined dynamically: whichever agent enters the corridor first forces the other to wait until it exits.

\paragraph{Waypoint Conversion}
We employ a string-pulling algorithm to convert classical MAPF guidance paths into waypoints for ORCA$^*$.
For each path $\pi^i \in \Pi$, the starting location is set as the first waypoint.
The algorithm iteratively identifies the longest prefix of the path that maintains line-of-sight from the previous waypoint; once visibility is obstructed by a static obstacle, the last visible vertex is selected as the next waypoint.
This process repeats until the goal location, which serves as the final waypoint.

The resulting waypoints guarantee that each segment between consecutive waypoints can be traversed in a straight line, ignoring other agents.
Such segments typically correspond to open areas or corridors.
Empirically, ORCA$^*$ performs effectively in open spaces for collision avoidance, while our spatial dependency mechanism handles corridors well. 
Thus, we found the waypoints generated by string-pulling are effective.

\subsection{ORCA$^*$-Execution}
\begin{algorithm}[t]
\caption{GetWaypoint($A,WP,\mathcal{A}$)}
\label{algo: getWP}
\small 
\begin{algorithmic}[1]
    \For{$i \gets A$}
        \If{$\boldsymbol{p}^i \approx$ currentWP($i$)}
            \State $WP'[i] \gets$ pop(WP[$i$])
        \EndIf
        \If{$\neg$LineOfSight($\boldsymbol{p}^i$, $WP'[i]$)}
            \State WP[$i$] $\gets$ newWP($i$, $WP'[i]$) $+$ WP[$i$]
            \State $WP'[i] \gets$ pop(WP[$i$])    
        \EndIf
        \If{$WP[i] \approx goal(i)$ \textbf{and} IsBlocking($i, \mathcal{A}(i)$)}
            \State $WP'[i] \gets$ YieldCell($i, \mathcal{A}(i)$)
        \EndIf
    \EndFor
    \State \Return $WP'$
\end{algorithmic}
\end{algorithm}

\begin{algorithm}[t]
\caption{CheckDependency($A,\mathcal{C},\mathcal{C}_{dep}$)}
\label{algo: depdency}
\small 
\begin{algorithmic}[1]
    \For{$i \gets A$}
        \If{approchingCorridor($i, \mathcal{C}$)}
            \If{hasAgent($c^i$,$A^{i_1}$) $||$ hasAgent($c^i$,$A^{i_2}$) }
                \State $c^i \gets c^i \cup \{\boldsymbol{p}^i\}$
                \State $mode(i) \gets drift$
            \EndIf
        \EndIf
        \If{ClearWait($i, c^i_{dep}$)}
            \State RemoveTempCells($i, c^i$)
            \State $mode(i) \gets ORCA$
        \EndIf
    \EndFor
\end{algorithmic}
\end{algorithm}

\begin{algorithm}[t]
\caption{PreferredVelocity($A, WP'$)}
\label{algo: prefV}
\small 
\begin{algorithmic}[1]
    \For{$i \gets A$}
        \If{$mode=ORCA$}
            \State $V^{pref}[i] \gets$ prefVORCA($i,WP'[i]$) 
        \EndIf
        
        \If{$mode=drift$}
            \If{hasNeighbour($i$)}
                \State $V^{pref}[i] \gets \max_{j \in \mathcal{A}(i)}$ prefVORCA($j,WP'[j]$)
            \Else
                \State $V^{pref}[i] \gets \alpha \cdot$ prefVORCA($i,\boldsymbol{p}^i_{start\_drifting}$) 
            \EndIf
        \EndIf
    
    \EndFor
    \State \Return $V^{pref}$
\end{algorithmic}
\end{algorithm}
The execution framework of C-ORCA$^*$ largely follows that of ORCA$^*$. Algorithm~\ref{algo: c-orca} outlines the high-level procedure, with highlighted lines indicating modifications from ORCA$^*$ and ORCA$^*$-MAPF.

At each time step, C-ORCA$^*$ first observes the environment and updates agent positions to account for execution uncertainty in practical settings.
Under the simulation setting, agents' positions are update assuming velocity are perfectly applied.
Agents are then grouped following the standard ORCA$^*$ procedure.
For each agent, the current waypoint is retrieved and updated if necessary, after which a preferred velocity is computed based on the agent’s mode and waypoint.
A collision-free velocity is subsequently determined using the same mechanism as vanilla ORCA$^*$, and finally, the selected velocities are applied to the agents.

In the following subsections, we describe the waypoint identification and dependency detection procedures and how they differ from ORCA$^*$.

\begin{figure}[t]
    \centering
    \begin{subfigure}[t]{0.3\columnwidth}
        \centering
        \begin{tikzpicture}[scale=0.43, every node/.style={scale=0.43}]

            \draw[thin, black] (0.5, 0.5) rectangle (6.5, 6.5);

            \fill[pattern=north east lines, pattern color=gray!70, draw=black] 
                (1.5, 1.0) -- (4.0, 1.0) -- (4.0, 4.7) -- 
                (2.8, 4.7) -- (2.8, 2.2) -- (1.5, 2.2) -- cycle;

            \draw[-{Stealth[length=1.0mm]},blue!70, thick, densely dashed] (3.7, 6.0) -- (6.1, 2.2);
            \draw[-{Stealth[length=1.0mm]},blue!70, thick, densely dashed] (2.4,4.4)-- (6.1, 2.2);

            \node[star, star points=5, star point ratio=2.25, draw=red!70, thick, inner sep=2pt, label=left:{$wp'$}] (wp_red1) at (2.4, 5.0) {};
            \node[star, star points=5, star point ratio=2.25, draw=red!70, thick, inner sep=2pt, label=right:{$wp''$}] (wp_red2) at (4.5, 5.0) {};
            \node[star, star points=5, star point ratio=2.25, draw=blue!70, thick, inner sep=2pt, label=below left:{$wp^*$}] (wp_blue) at (6.2, 2.0) {};

            \node[circle, fill=blue!70, draw=black, minimum size=0.5cm, inner sep=0pt, label={[xshift=2pt, yshift=-2pt]right:{$a^1$}}] (a_prime) at (3.7, 6.0) {};
            \node[circle, fill=blue!70, draw=black, minimum size=0.5cm, inner sep=0pt, label=left:{$a^{1'}$}] (a) at (2.4, 4.4) {};

            \draw[-{Stealth[length=2.5mm]}, blue!70, thick] (a_prime) -- (a);
            
        \end{tikzpicture}
        \caption{Line of sight interruption}
        \label{fig:scenario_los}
    \end{subfigure}
    \hfill
    \begin{subfigure}[t]{0.3\columnwidth}
        \centering
        
        \begin{tikzpicture}[scale=0.43, every node/.style={scale=0.43}]
             
            \draw[thin, black] (0.5, 0.5) rectangle (6.5, 6.5);
            
            
            \filldraw[pattern=north east lines, pattern color=gray!70] (3,3) rectangle (4,6);
            

            
            
            \node[star,star points=5,star point ratio=2.25, scale=0.5,
                  draw=red!80,fill=white,minimum size=8pt,thick]
                  (wpp22) at (2,2.5) {};
            \node[red!80,below=2pt] at (2.3,2.6) {$\mathrm{wp}_2^{''}$};
            
            \node[circle, fill=blue!70, draw=black, minimum size=0.5cm, inner sep=0pt, label={[xshift=2pt, yshift=-2pt]right:{$a^1(g^1)$}}] (a_prime) at (4.5, 2.5) {};
            \node[circle, fill=red!70, draw=black, minimum size=0.5cm, inner  sep=0pt, label=above:$a^2$] (a2) at (5.5, 3.5) {};

            \node[star,star points=5,star point ratio=2.25, scale=0.5, draw=red!80,fill=blue,minimum size=8pt,thick] (wp1) at (4.5,2.5) {};
            \node[red!80,below=2pt] at (4.5,2.5) {$\mathrm{wp}_2^{'}$};
            
            \draw[red!80,dashed,thick, ->,  >=stealth] (4,2.5) -- (2.4,2.5);
            \draw[red!80,dashed,thick, <-,  >=stealth] (4.65,2.75) -- (5.3,3.3);
            
        \end{tikzpicture}
        
        \caption{Local waypoint generation}
        \label{fig:scenario_wp}
    \end{subfigure}
    \hfill
\begin{subfigure}[t]{0.3\columnwidth}
        \centering
        \begin{tikzpicture}[scale=0.43, every node/.style={scale=0.43}]
            
            \draw[thin, black] (0.5, 0.5) rectangle (6.5, 6.5);
            \fill[pattern=north east lines, pattern color=gray!70, draw=black] (1.5, 4.0) rectangle (5.5, 5.0);
            \fill[pattern=north east lines, pattern color=gray!70, draw=black] (1.5, 2.0) rectangle (5.5, 3.0);
            \node[circle, fill=red!70, draw=black, minimum size=0.5cm, inner  sep=0pt, label=below:$a^2$] (a2) at (1.0, 2.2) {};
            \draw[red!70, dashed, thick, -{Stealth[length=2.5mm]}] (a2) to[out=90, in=180] (2.2, 3.5);
            \node[circle, draw=black,  fill=blue!70, minimum size=0.5cm, inner sep=0pt, label=right:$a^1$] (a1) at (4.3, 3.5) {};
            \draw[blue!70, thick, -{Stealth[length=2.5mm]}] (a1) -- (2.5, 3.5);
            \node[circle, draw=black, fill=blue!70, minimum size=0.5cm, inner sep=0pt, label=above:{$a^3$}] (a3) at (6.0, 3.9) {};
            \draw[blue!70, thick, -{Stealth[length=2.5mm]}] (a3) to[out=250, in=0] (5.0, 3.4);
            
        \end{tikzpicture}
        \caption{Multiple agent trailing behavior}
        \label{fig:scenario_trailing}
    \end{subfigure}
    
    \caption{Different waypoint updates and corridor trailing situations}
    \label{fig:all_scenarios}
\end{figure}
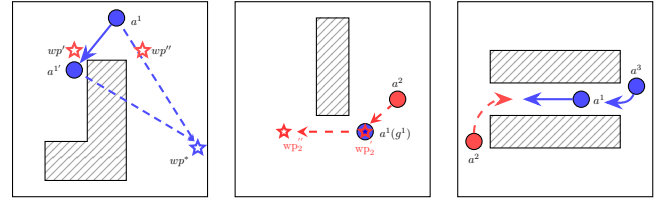
\paragraph{Get Waypoint}
Shown in Algorithm~\ref{algo: getWP}, every agent retrieves its current target waypoint and updates it if necessary at each timestep.
$WP$ denotes the full waypoint sequence for each agent, and $WP'$ is the current waypoint.

If agent $a^i$ has reached its current waypoint $WP'[i]$, the next waypoint is popped from $WP[i]$ to replace it (lines 2–4).
If the line-of-sight between the agent’s current position and $WP'[i]$ is obstructed, intermediate waypoints are generated (lines 5–8) via string-pulling on a single-agent grid path from the current position to the waypoint.
These waypoints ensure line-of-sight between consecutive segments; a violation during execution indicates that the agent has deviated from its intended path.
Maintaining line-of-sight is critical, as ORCA$^*$’s myopic nature can otherwise lead to deadlocks, even without external interference.
Fig.~\ref{fig:scenario_los} illustrates such a case: agent $a^1$ may be forced towards the left side of the obstacle, $a^{1'}$, by other agents. 
$a^1$ will deadlock by itself in this configuration as the locally optimal behaviour is to move down into the alcove. Navigating around the obstacle requires temporarily moving away from the waypoint.
To address this issue, two intermediate waypoints ${wp}'$ and ${wp''}$ are introduced to guide the agent out of the position.

If agent $a^i$ has reached its final goal $g^i$ but blocking agent $a^j$'s next waypoint $wp'_j$, i.e. $wp'_j=g^i$, a temporary waypoint at $a^j$’s current position $\boldsymbol{p}^j$ is assigned to $a^i$ as a yielding position (lines 9–11).
This is because an agent at its goal has a preferred velocity of $(0,0)$, so the locally optimal resolution between $a^i$ and $a^j$ may cause them to deadlock.
By assigning a temporary waypoint, we introduce a non-zero preferred velocity, breaking this local optimum.
This is mechanism illustrated in Fig.~\ref{fig:scenario_wp}. Agent $a^1$ is stationed at its goal $g^1$, which doubles as the current waypoint of $a^2$.
Assigning $a^1$ a temporary waypoint at $\boldsymbol{p}^2$ encourages it to vacate the blocking position while simultaneously inducing a non-zero preferred velocity for $a^1$, thereby resolving the deadlock.

\paragraph{Dependency Detection}
Algorithm~\ref{algo: depdency} outlines the procedure for handling dependencies during execution, where $\mathcal{C}$ denotes all the corridors and $\mathcal{C}_{dep}$ denotes all the dependency groups corresponding to $\mathcal{C}$.

Agents can be in either \textit{normal} or \textit{drift} mode.  When an agent $a^i$ approaches a corridor $c^i$ that is currently occupied by agents belonging to one of the precomputed dependency groups $A_{i_1}$ or $A_{i_2}$ (lines 2–3),
its current position $\boldsymbol{p}^i$ is treated as part of a temporarily expanded corridor, and switches to drift mode (lines 4–5). This temporary expansion prevents excessive inflow of agents into the corridor, mitigating congestion effects.
For instance, in cross-corridor scenarios, agents waiting at the entrance may be pushed into the corridor by the swarm behind, potentially blocking agents inside from exiting.
Agents in drift mode move adaptively with their local neighbourhood without a fixed preferred velocity---effectively allowing agents inside the corridor to push their way out---as opposed to waiting (i.e. a preferred velocity set to $(0,0)$). 
The mechanism of drifting agents is detailed below.

The $ClearWait$ procedure (line 8) determines whether a drifting agent can resume normal execution based on the dependency type.
For example, consider an agent $a^i \in A_{i_1}$ subject to a cross-corridor dependency for corridor $c^i$.
If no agent $a^j \in A_{i_2}$ remains in $c^i$, then $a^i$ is released from waiting.
Consequently, all temporary corridor expansions are removed, and drifting agents revert to the standard ORCA$^*$ mode.

Under this dependency scheme, suppose $a^2 \in A_{i_2}$ is waiting for $a^1 \in A_{i_1}$ to exit the corridor, as illustrated in Fig.~\ref{fig:scenario_trailing}.
If additional agents from $A_{i_1}$, such as $a^3$, enter the corridor while $a^1$ is still traversing, then $a^2$ will defer entry until all agents in $A_{i_1}$ have cleared the corridor.

\paragraph{Get Preferred Velocity}
Algorithm~\ref{algo: prefV} outlines the computation of preferred velocities under different agent modes.

For agents in the standard ORCA$^*$ mode, the preferred velocity is computed identically to vanilla ORCA$^*$ (lines 2–4).
For agents in drifting mode, if neighbouring agents are present, the preferred velocity is set to that of the neighbour with the largest magnitude (lines 6–7).
Otherwise, the agent gradually returns toward its drifting origin (line 9), scaled by a factor $\alpha \in (0,1)$.
Agents with reduced velocity are easier to push by nearby agents, which is desirable for drifting behaviour.

Note that ORCA$^*$-MAPF employs a fallback mechanism to resolve deadlocks after they occur.
Our approach is orthogonal and aims to prevent deadlocks proactively; however, residual deadlock cases may still arise.
So, ORCA$^*$-MAPF remains applicable to C-ORCA$^*$ and the two methods can be naturally combined.

\section{Experiment}
C-ORCA$^*$ and C-ORCA$^*$-MAPF \footnote{https://git.andyli.info/andyli/C-ORCA}was implemented on top of the existing ORCA$^*$-MAPF implementation \footnote{https://github.com/PathPlanning/ORCA-algorithm}.
For the {-MAPF} variant, a deadlock threshold of 250 time steps is used, i.e. the -MAPF fall-back mechanism will be triggered if agents are not moving for 250 time steps.
All the benchmarks are evaluated on a server equipped with an AMD EPYC-Milan processor with 64 cores and 128GB of RAM.
The results are compared against ORCA$^*$ and ORCA$^*$-MAPF on three main metrics: success-rate, runtime and flowtime cost. 
The simulation terminates if one of the three following conditions is met:
\begin{enumerate}
    \item all agents reach their goals; this is the only success condition
    \item the mean velocity falls below 0.0001 over a sliding window of 1,000 time steps
    \item the simulation exceeds 30,000 time steps
\end{enumerate}

\subsection{Benchmarking}
Each agent was modelled as a disk with a true radius of 0.3 units, with an additional safety margin of 0.1 units per agent during the computation of feasible velocity for agents to reduce the risk of collisions. Agents have a maximum speed of 1 unit per second simulated at 10 steps per second, with a local observation range of 3 units.
Four different types of maps were used:
\begin{itemize}
    \item \textbf{Gap} ($64 \times 64$): Two large areas separated by a wall with a unit wide passage. Agents were equally divided between the two areas with goals on the opposite side.
    \item \textbf{Random} ($64 \times 64$): Randomly distributed obstacles. Start and goal locations were randomly assigned.
    \item \textbf{Room} ($32 \times 32$): Rooms with interconnected doors. Start and goal locations were randomly assigned.
    \item \textbf{Warehouse} ($321 \times 123$): Long, parallel narrow aisles with 1 agent size width. Start and goal locations were randomly assigned.
\end{itemize}
Gap was included as it was the most challenging map reported for ORCA$^*$-MAPF. Random, Room and Warehouse were sourced from \cite{MAPF_Tracker}.
The number of agents varied from 10 to 200 in increments of 5; 50 problem instances were run for each number of agents.
For C-ORCA$^*$, we used MAPF-LNS to compute the initial guidance path. 
\begin{figure*}[t]
    \centering
    \includegraphics[width=0.95\textwidth]{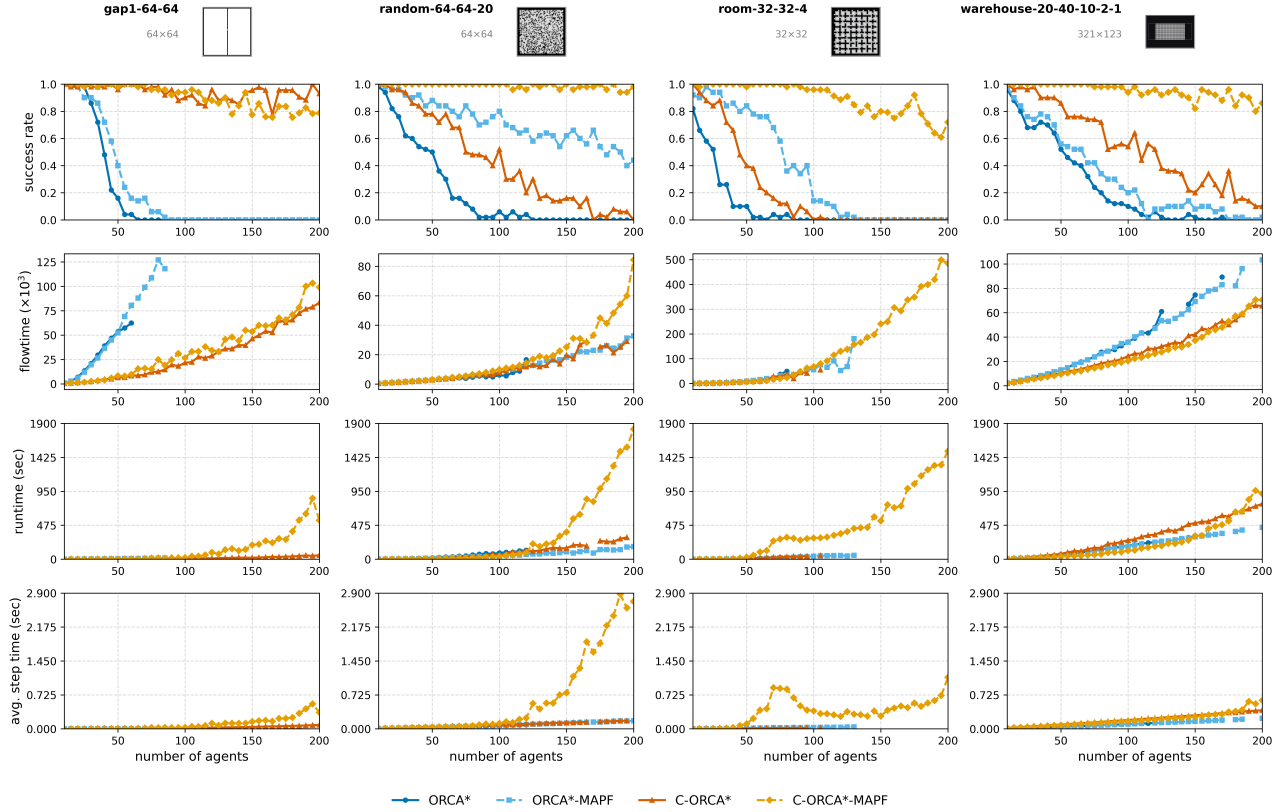}
    \caption{Comparison of four algorithms across four map types. Each column corresponds to a different map. The rows show, from top to bottom, the success rate,  flowtime ($\times 10^3$), runtime (sec), and average step time (sec) as a function of the number of agents. Failed instances are not included in the plots.}
    \label{fig:experiment}
\end{figure*}

\begin{figure*}[h]
    \centering
    \includegraphics[width=0.95\textwidth]{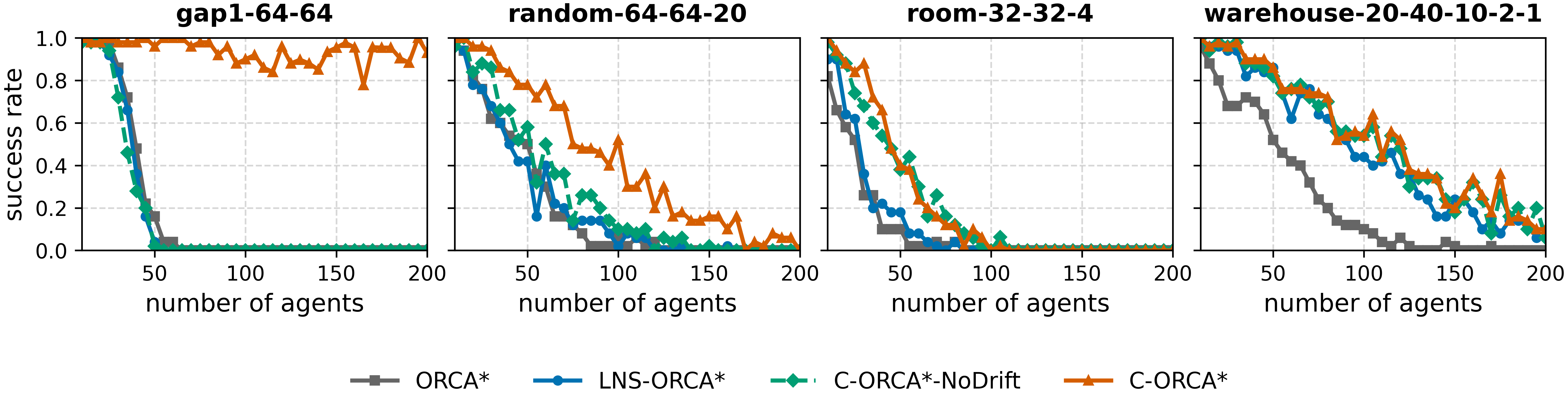}
    \caption{Success rate for ablation experiment}
    \label{fig:ablation}
\end{figure*}

\begin{table*}[t]
  \centering
  \resizebox{\textwidth}{!}{
  \begin{tabular}{ccccccccc}
    \toprule
    \multirow{2}{*}{Agents} & \multicolumn{2}{c}{Gap} & \multicolumn{2}{c}{Random} & \multicolumn{2}{c}{Room} & \multicolumn{2}{c}{Warehouse} \\
    \cmidrule(lr){2-3} \cmidrule(lr){4-5} \cmidrule(lr){6-7} \cmidrule(lr){8-9}
     & ORCA$^*$-MAPF & C-ORCA$^*$-MAPF & ORCA$^*$-MAPF & C-ORCA$^*$-MAPF & ORCA$^*$-MAPF & C-ORCA$^*$-MAPF & ORCA$^*$-MAPF & C-ORCA$^*$-MAPF \\
    \midrule
    10 & 0.06 & 0.0 & 0.12 & 0.0 & 1.85 & 0.19 & 0.2 & 0.0 \\
    15 & 1.16 & 0.0 & 0.33 & 0.1 & 3.18 & 0.51 & 0.42 & 0.04 \\
    35 & 16.74 & 0.02 & 6.51 & 0.38 & 51.79 & 9.93 & 54.31 & 0.48 \\
    55 & 65.75 & 0.06 & 24.02 & 1.11 & 73.72 & 5.35 & 216.44 & 0.91 \\
    75 & 5.67 & 0.1 & 63.76 & 2.05 & 629.17 & 15.84 & 1309.38 & 2.94 \\
    95 & - & 0.19 & 115.42 & 2.56 & 448.75 & 17.48 & 2191.83 & 2.5 \\
    115 & - & 0.16 & 190.53 & 3.4 & 544.2 & 22.75 & - & 3.33 \\
    135 & - & 0.28 & 278.72 & 3.63 & - & 10.53 & 1201.4 & 73.42 \\
    155 & - & 0.31 & 461.27 & 5.22 & - & 17.2 & 923.2 & 53.03 \\
    175 & - & 0.2 & 482.93 & 8.36 & - & 28.5 & - & 40.81 \\
    195 & - & 0.39 & 566.65 & 5.03 & - & - & - & 126.53 \\
    200 & - & 0.22 & 651.45 & 4.49 & - & - & 1513.0 & 364.55 \\
    \bottomrule
  \end{tabular}
  }

  \caption{Average number of MAPF calls per instance.}
  \label{tab:mapf_calls}
\end{table*}


\subsection{Results}

Figure~\ref{fig:experiment} plots success rate, flowtime, run time, and average step time across various maps as a function of agent count. Table~\ref{tab:mapf_calls} shows the average number of MAPF calls across instances. 
Overall, C-ORCA$^*$-MAPF consistently outperformed all other algorithms in success rate, runtime, and flowtime across all four maps. 

\textbf{Success rate.} Figure~\ref{fig:experiment}’s topmost row shows that C-ORCA$^*$-MAPF achieves the highest success rate across almost all maps,
with the exception of \textit{gap1-64-64}, where its performance is approximately the same as C-ORCA$^*$.
C-ORCA$^*$ consistently out performs ORCA$^*$, even ORCA$^*$-MAPF for Gap and Warehouse. 
On Gap, C-ORCA$^*$ maintains a success rate close to 1.0 for all tested agent counts, while ORCA$^*$ drops below 0.4 at 30 agents and fails almost entirely beyond 75.
Gap forces agents to traverse narrow passages from opposite directions, inevitably leading to head-on tension that ORCA$^*$ cannot resolve.
On Random, C-ORCA$^*$-MAPF sustains a success rate around 1 even with 200 agents, whereas ORCA$^*$-MAPF has a success rate of around 0.5.
The algorithms generally performed poorly on Room. One critical reason is that this is the smallest map, agents are most densely packed. At 200 agents 30\% of traversable tiles are occupied. Yet, C-ORCA$^*$-MAPF is still capable of solving almost all instances until around 100 agents, and the success rate gradually drops to 60\% at 200 agents.
Warehouse is composed of almost entirely corridors, and head-on encounters were frequent.
The success rate of ORCA$^*$ and ORCA$^*$-MAPF degraded rapidly as agent count increased, while C-ORCA* and C-ORCA$^*$-MAPF degraded much slower.
C-ORCA$^*$-MAPF remains successful above 80\% of the time with 200 agents.
The substantial improvement of C-ORCA$^*$-MAPF over ORCA$^*$-MAPF is primarily due to the main failure mode of ORCA$^*$-MAPF.
In ORCA$^*$-MAPF, agents often become densely packed together, preventing them from reaching their designated starting vertices for MAPF mode.
As a result, the system fails to initiate MAPF mode altogether.
In contrast, the proactive waiting mechanism in C-ORCA$^*$-MAPF maintains larger inter-agent distances even when deadlocks occur.
This additional spacing allows agents to successfully reach their designated starting vertices, thereby enabling MAPF mode to start successfully.

\textbf{Ablation.}
We present the following ablation studies: (1) ORCA augmented with only the LNS MAPF guidance path (LNS-ORCA*), and (2) ORCA augmented with corridor dependencies but without drift mode (C-ORCA$^*$-NoDrift). These are compared against the full C-ORCA$^*$ implementation. The results are shown in Fig.~\ref{fig:ablation}.

Results indicate that the impact of drift mode is most pronounced on Gap, where the guidance path contributes little to none.
Corridor dependencies alone are insufficient in this setting, as agents become trapped in the corridor.
Consequently, drift-enabled C-ORCA$^*$ maintains a near-100\% success rate, whereas C-ORCA$^*$-NoDrift performs as poorly as ORCA*.
On Random, C-ORCA$^*$-NoDrift provides only marginal improvements over ORCA*, while the full C-ORCA* achieves substantially higher success rates for similar reason.
On Room, the guidance path appears to provide limited benefit, likely because the generated guidance paths closely resemble the agents' individual shortest paths.
Furthermore, agents are distributed across multiple corridors, reducing congestion and allowing corridor dependencies alone to provide sufficient coordination.
As a result, drift mode yields only a modest additional improvement. In contrast, the guidance path plays a more significant role on Warehouse, where deciding which corridor to go through is critical.
Corridor dependencies and drift mode provide little additional improvement in warehouse maps, as agents interact far less once corridor usage becomes evenly distributed.

\textbf{Flowtime.} The second row of Figure~\ref{fig:experiment} reflects the overall solution cost.
C-ORCA$^*$ and C-ORCA$^*$-MAPF consistently achieved lower flowtime than their ORCA$^*$ counterparts for co-solvable instances, cost for C-ORCA$^*$-MAPF seems higher than other algorithms because it has much higher success rate on harder instances.
The difference is most pronounced on the Gap and Warehouse map, where C-ORCA$^*$(-MAPF) is able to consistently solve instances  with 50\%  or more of the ORCA$^*$(-MAPF) cost on these two maps.
We interpret the lowered flowtime as the magnitude in which our dependency mechanisms improve solution quality. Agents in ORCA$^*$ frequently become trapped near narrow passages, wasting time before the deadlock gets resolved or timing out.

\textbf{Runtime. \& average step time.} Runtime (row 4) shows the overall runtime of the algorithm, excluding the time to compute the initial guidance path. Average step time (row 5) is the average computation time per time step.
-MAPF variants scale very poorly with the number of agents in general; this is due to the fact that the fallback solver is using ECBS~\cite{ECBS}, which fail with higher numbers of agents.
We can see that the dependency mechanism almost introduces no time overhead, as the average time per time-step for C-ORCA$^*$ is almost identical to ORCA$^*$.

\textbf{MAPF calls.} Table~\ref{tab:mapf_calls} records the number of times that MAPF fall-back mechanisms are invoked.
This reflects the performance of the dependency mechanism, as each MAPF fallback call is expensive in terms of both computation time and solution quality.
A MAPF fallback call also implies that a solution cost of least 25 $\times$ number of agents was wasted, where 25 is the deadlock threshold (250 timesteps) $\times$ cost per time step (0.1).
As mentioned earlier, the MAPF fall-back call is ECBS, which is generally expensive in a real-time planning setting.
From Table~\ref{tab:mapf_calls}, it’s clear that although C-ORCA$^*$-MAPF still invokes MAPF, it does so on up to 3–4 orders of magnitude less frequently.

\section{Conclusion}
In this paper we extend ORCA$^*$ to C-ORCA$^*$ by guiding the agents with a classical MAPF planning algorithm, detecting dynamic dependencies, and introducing mechanisms to proactively resolve challenging deadlock scenarios during the execution.
C-ORCA$^*$ (resp. -MAPF) consistently outperforms ORCA$^*$ (resp. -MAPF), and achieves a significantly higher success rate and solution quality than previous state-of-the-art across all tested benchmarks, and handles scenarios that ORCA$^*$ previously struggled with. However, under extreme congestion, the algorithm may converge to a local optimum.
Future work could incorporate a mechanism based on sampling-based search, with a gradual transition to systematic search within this framework, to further improve the algorithm’s performance. Nevertheless, the C-ORCA$^*$ family of algorithms represents a significant step towards realising the practical control of large robot fleets.

\bibliography{aaai2026}
\end{document}